\def\year{2022}\relax
\documentclass[letterpaper]{article} % DO NOT CHANGE THIS
\usepackage{aaai22}  % DO NOT CHANGE THIS
\usepackage{times}  % DO NOT CHANGE THIS
\usepackage{helvet}  % DO NOT CHANGE THIS
\usepackage{courier}  % DO NOT CHANGE THIS
\usepackage[hyphens]{url}  % DO NOT CHANGE THIS
\usepackage{graphicx} % DO NOT CHANGE THIS
\usepackage{natbib}  % DO NOT CHANGE THIS AND DO NOT ADD ANY OPTIONS TO IT
\usepackage{caption} % DO NOT CHANGE THIS AND DO NOT ADD ANY OPTIONS TO IT
\DeclareCaptionStyle{ruled}{labelfont=normalfont,labelsep=colon,strut=off} % DO NOT CHANGE THIS
\usepackage{algorithm}
\usepackage{algorithmic}
\usepackage{booktabs}
\usepackage{amsmath}

\usepackage{newfloat}
\usepackage{listings}
\floatstyle{ruled}
\newfloat{listing}{tb}{lst}{}
\floatname{listing}{Listing}
\usepackage{soul, color, xcolor}

\newcommand{\methodname}[1]{FedNLG}
\newcommand{\seqseq}[1]{SEQ2SEQ}
\newcommand{\persona}[1]{Persona}
\newcommand{\friend}[1]{Friends}
\newcommand{\tbbt}[1]{The Big Bang Theory}
\newcommand{\cornell}[1]{Cornell Movie-Dialogs}

\title{Federated Natural Language Generation for Personalized Dialogue System}
\author{
    Yujie Lu\equalcontrib, \textsuperscript{\rm 1}
    Chao Huang\equalcontrib, \textsuperscript{\rm 2}
    Huanli Zhan, \textsuperscript{\rm 2}
    Yong Zhuang \textsuperscript{\rm 3}
}
\affiliations{
    \textsuperscript{\rm 1} University of California, Santa Barbara\\
    \textsuperscript{\rm 2} Tencent\\
    \textsuperscript{\rm 3} Carnegie Mellon University\\
    yujielu@ucsb.edu, codyhuang@tencent.com, sidzhan@tencent.com, yongzhua@andrew.cmu.edu
}

\usepackage{bibentry}
\begin{document}

\maketitle

\begin{abstract}
% (or large noisy open-domain dataset, OpenSubtitles)
Neural conversational models have long suffered from the problem of inconsistency and lacking coherent personality.
To address the issue, persona-based models capturing individual characteristics have been proposed, but they still face the dilemma of model adaption and data privacy.
To break this dilemma, we propose a novel Federated Natural Language Generation (\methodname~) framework, which learns personalized representations from various dataset on distributed devices, and thus implements the personalized dialogue system efficiently and safely.
%\methodname~ first pre-trains parameters of standard neural conversational model over \cornell~ Corpus, and then fine-tune the model parameters and persona embeddings on \tbbt~ and \friend~ datasets, in a federated manner.
\methodname~ first pre-trains parameters of standard neural conversational model over a large dialogue corpus, and then fine-tune the model parameters and persona embeddings on specific datasets, in a federated manner.
Thus, the model could simultaneously learn the persona embeddings in local clients and learn shared model parameters by federated aggregation, which achieves accuracy-privacy balance.
By conducting extensive experiments, we demonstrate the effectiveness of our model by pre-training model over \cornell~ Corpus and fine-tuning the model over two TV series dataset.
%demonstrate the effectiveness of our model by first 
%With the proposed approach, we achieve XX improvement on metrics of BLEU scores and Perplexity.
% \yj{may add percent performance}
\end{abstract}

\section{Introduction}

% introduce conversational agents
As the widely application of conversational agents in digital world, such as house-hold robot, speech assistant, intelligent customer services, there has been growing research interest in training naturalistic conversation systems from large volumes of human-to-human interactions. 
In addition, there is an increasing need for personalized service, as personalization enables clients feel incorporated as an integral part of user experience.

\begin{figure}[h]
    \centering
    \includegraphics[width=.9\columnwidth]{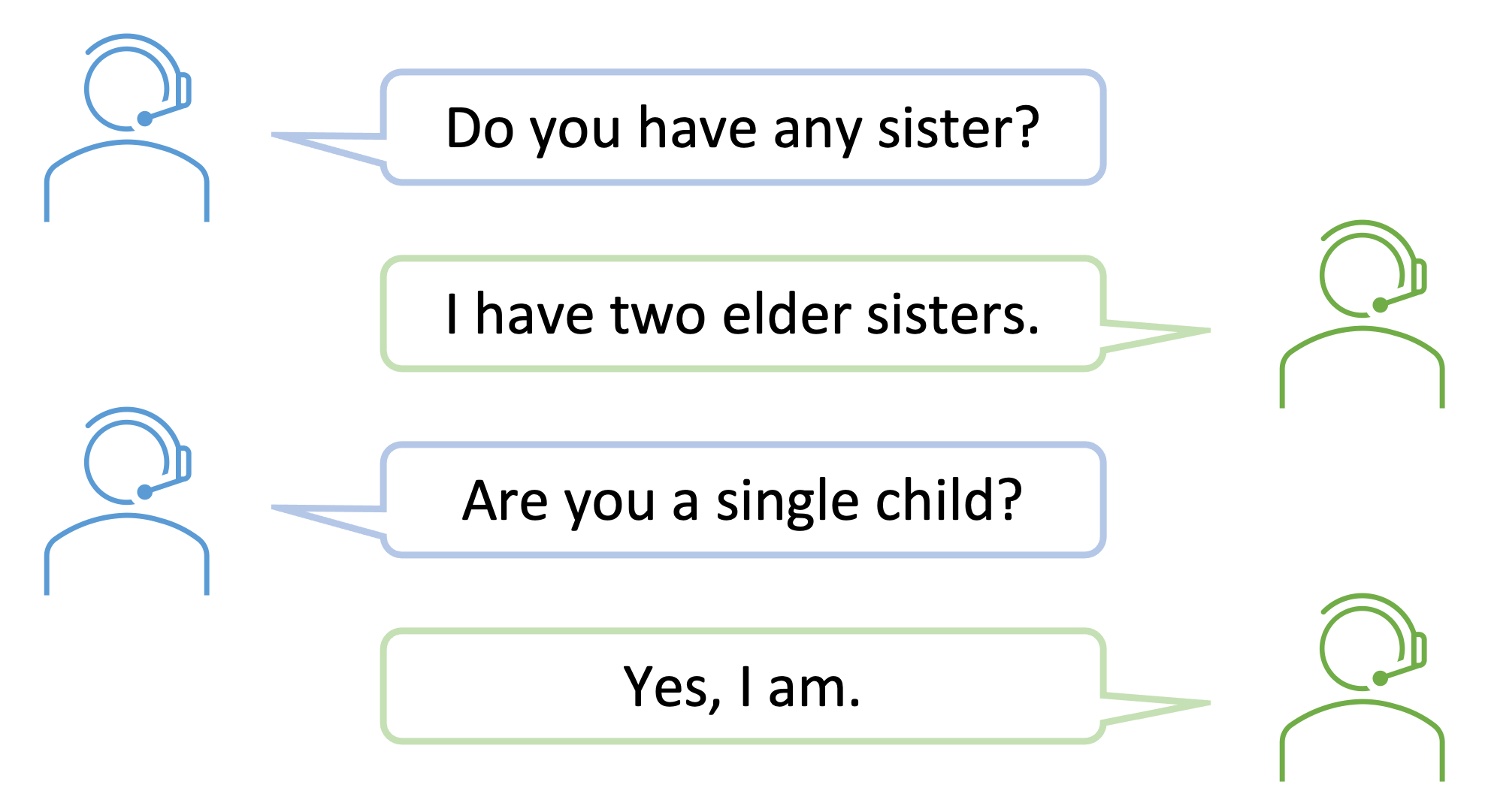}
    \caption{Predicting inconsistent responses with little humanity is a common issue in conversational models.}
    % based on sequence-to-sequence \seqseq~ framework.
    \label{fig:problem1}
\end{figure}

\begin{figure}[h]
    \centering
    \includegraphics[width=.8\columnwidth]{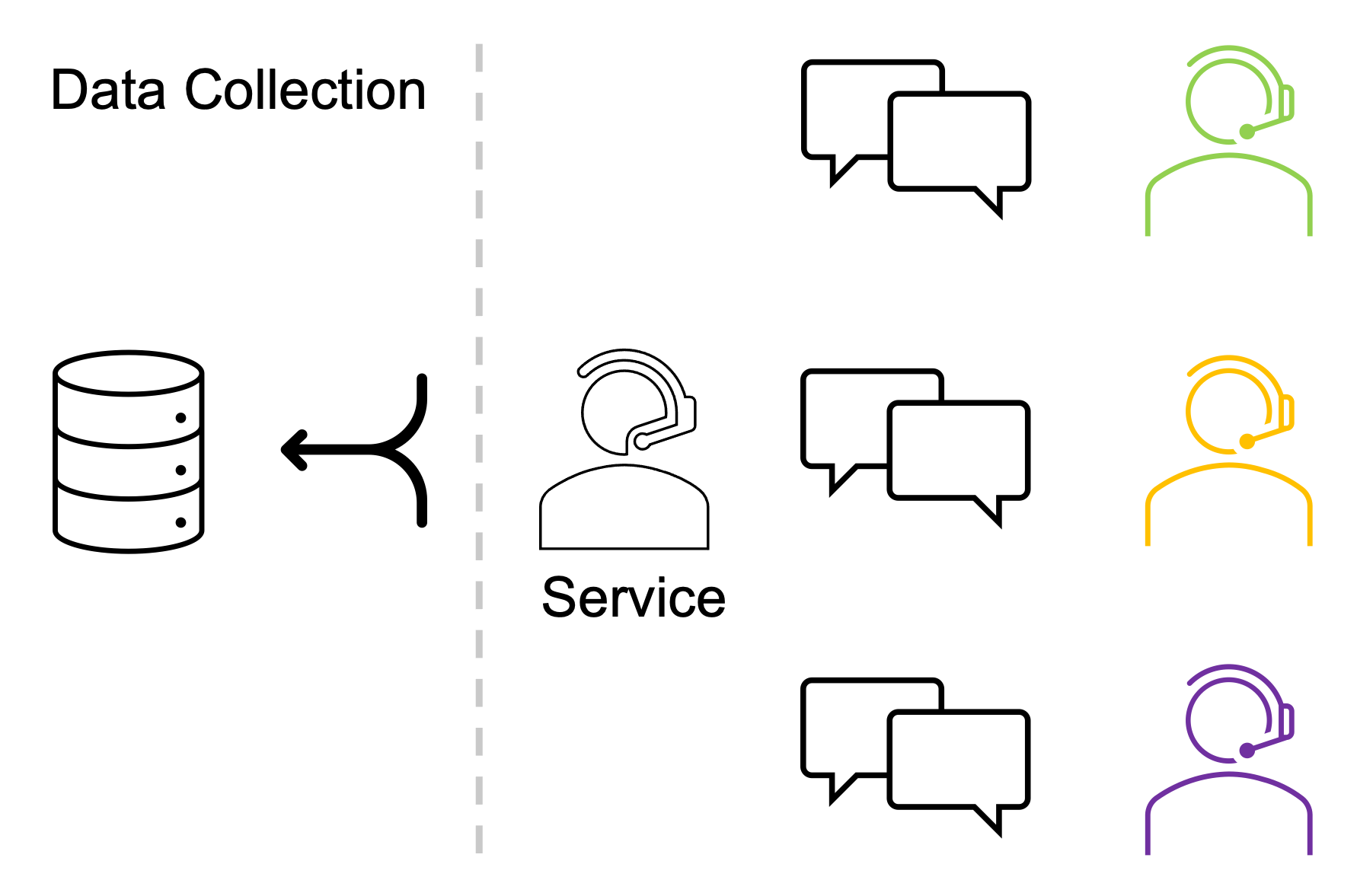}
    \caption{Existing works utilize rich data to boost the model accuracy while sacrifice data privacy due to their need to collect personal information and chat history.}
    \label{fig:problem2}
\end{figure}

Existing methods to generate dialogue responses are mainly based on neural machines.
However, such corpus-driven methods argue for very large corpora and corpus evidence are not fully exploited, as discussed in \cite{richard2009corpus}.
Besides, such methods utilize data itself as the sole source of hypotheses \cite{Hinkka2018DatadrivenLT}, resulting in conversational models with low perplexity.
We illustrate the issue of producing responses with consistency and humanity in Figure~\ref{fig:problem1}.
Although recent work like \cite{LiPersona} addresses the challenge of inconsistency by successfully endowing data-driven systems with the coherent ``persona" to model human-like behavior, neural machine based models still face a data privacy issue.
As shown in Figure~\ref{fig:problem2}, current persona-based models directly uses all conversational data from users during the training process, which results in a data privacy issue.

% generating vague and inconsistent responses with low perplexity.
% Thus can be widely inconsistent and lack of personalization.
% the propensity to select the response with the greatest likelihood is a major issue.

% introduce two challengs
%As shown in Figure~\ref{fig:problem1} and Figure~\ref{fig:problem2}, there are two major challenges in this scenario: 1) producing responses with consistency and humanity; and 2)
%achieving balance between model accuracy and data privacy.

%Previous neural machine based models, though achieve good scores in BLEU and Perplexity, still fail these tow major challenges.
% perform poorly in human-evaluation for these two challenges, though achieve good scores in BLEU and Perplexity.

% There are some... Existing works successfully address the inconsistency issue via persona-based neural conversational models but suffer from the data privacy issue due to their need to collect personal information and dialogue history for centralized training.

\begin{figure}
    \centering
    \includegraphics[width=.8\columnwidth]{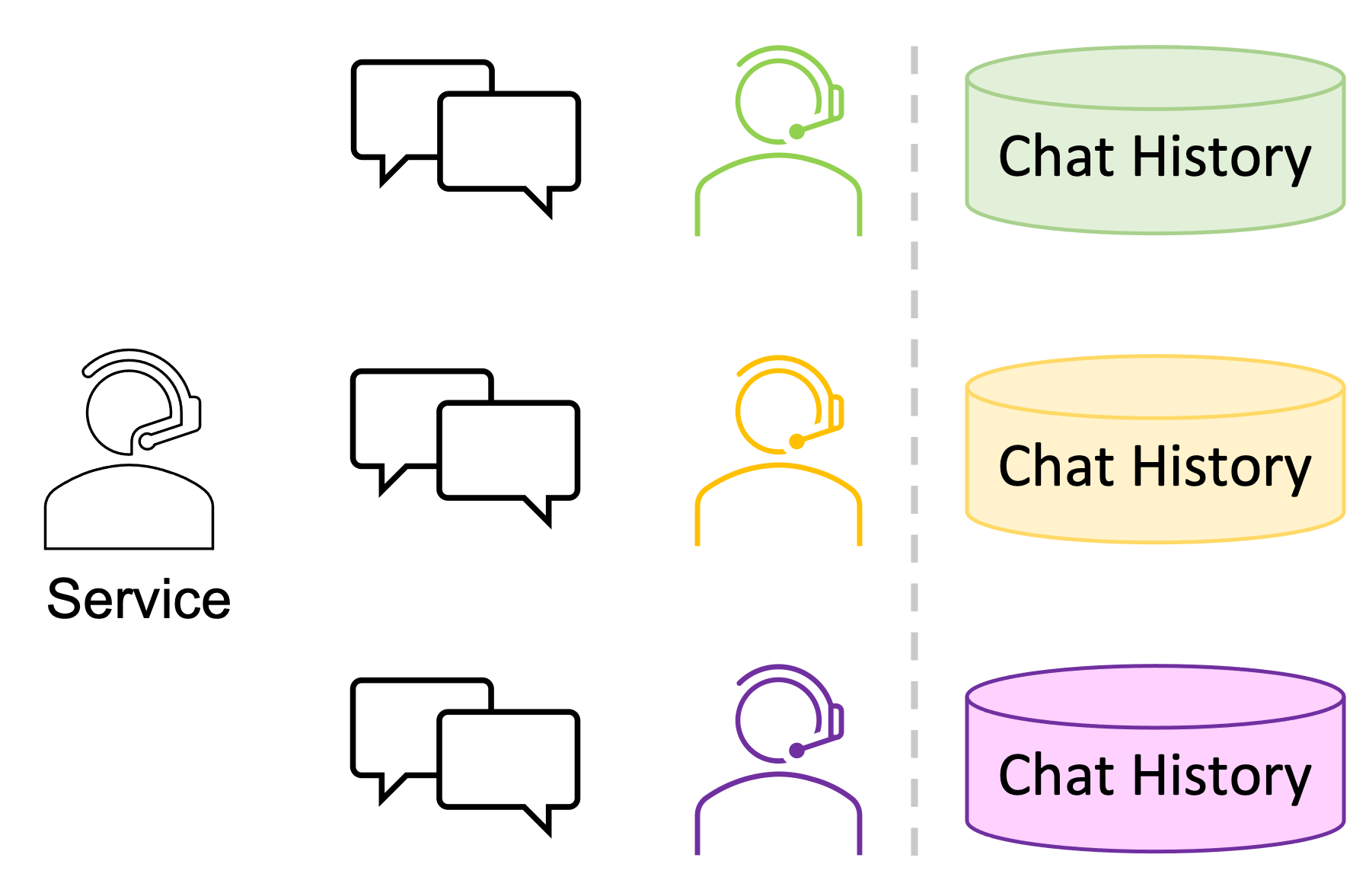}
    \caption{Our proposed framework is capable of provide personalized service with humanity, while keep the chat history locally and achieves accuracy-privacy balance.}
    \label{fig:solution}
\end{figure}

% our "implementation"
In this paper, we propose a framework, Federated Natural Language Generation (\methodname~), to address the challenge of consistency with humanity and accuracy-privacy balance at the same time.
To tackle the challenge of consistency with humanity, we first embed user personal profile information (e.g., age, gender, location, education), and historical user behaviors (e.g., dialogue history, application using time duration)  into profile embedding and behavior embedding, respectively.
Then, we concatenate both of the embeddings into one single persona-based embedding.
To tackle the challenge of data privacy, Federated Learning (FL) is a solution, which is a technique where machine learning models are trained in a decentralized manner.
In particular, instead of explicitly exchanging client data, FL exchanges model parameters to avoid the data privacy issue.
% Federated learning for dialogue system was previously explored in \cite{joel2020fednext}, in which a federated recurrent neural network (RNN) was trained for next-word prediction.
%Thus, we learn the general embedding and persona embedding among clients by \methodname~ framework, which is designed for privacy-preserving natural language generation.
Thus, we learn the general embedding and persona embedding among clients by FL, which is designed for privacy-preserving natural language generation.
As in Figure~\ref{fig:solution}, the \methodname~ framework perfectly combines the persona-based dialogue system with federated learning.

Specifically, in this work, we first pre-train a general conversational agent within a sequence-to-sequence (\seqseq~) framework over \cornell~ Corpus dataset.
Then we finetune and incorporate persona embeddings over \tbbt~ and \friend~ movie scripts. During fine-tuning, we adopt our proposed \methodname~ framework, which represents each persona as a local client, and each client upload the local parameters to the central server, and download the globally aggregated parameters from the central server to update the local model.
The aggregation is adapt to existing commonly used federated techniques, such as FedAvg, which takes the local model parameters from each client as input, and output the averaged model parameters to each client for update without personal data shared.
%Besides, the model parameters is split into private parameters, which relates to persona, and federated parameters, which relates to sentence understanding.
%The persona embeddings are trained on personal conversational data and used at test time to generate personalized consistent responses with certain style and back information.

In this paper, our contributions are as follows:
\begin{itemize}
\item We propose \methodname~, a privacy-preserving persona-based framework that enables safe personalized services in the domain of natural language generation.
% \item We devise the parameter separation technique to successfully incorporate federated learning into personalized neural conversational models.
% and improve relative performance than standard \seqseq~ model up to ${xx\%}$ and ${xx\%}$ in BLEU scores, ${xx\%}$ and ${xx\%}$ in perplexity, respectively.
% \methodname~ shows superiority of BLEU scores and perplexity over baselines in both \tbbt~ and \friend~ dataset.
% \item We investigate the parameter separation for federated learning. We analyze the influence of different federated learning configurations in our proposed framework and empirically show that FedProx achieve state-of-the-art performance in federated personalized natural language generation.
\item We empirically show that the \methodname~ framework significantly improves the performance of models in the federated personalized setting and conclude that FedProx is more appropriate in federated personalized natural language generation. 
% Our experiments over \tbbt~ and \friend~ dataset show that leveraging federated persona embeddings can achieve similar results with persona-based models while preserving data privacy.
\item Ablation study show that our proposed \methodname~ is flexible and could be applied to exisiting widely used federated algorithms, including FedAvg and FedProx, etc.
\end{itemize}

\section{Related Works}
\subsection{Neural Conversational Modeling}
Conversational modeling is a challenging research topic as it requires complex mapping between queries and responses. Previous works are restricted to specific domains and need feature engineering.
Inspired by recent works \cite{kal2013recurrent, Sutskever2014SequenceTS, Bahdanau2015NeuralMT} which map sequences to sequences with neural networks, \cite{Vinyals2015neural} proposed a neural conversational model based on \seqseq~ framework, which are capable of generating fluent and accurate replies in conversation in an end-to-end manner.

A common problem is that the produced responses often lack persona and consistency. Personalized responses are essential for providing an informative and human-like conversation. Thus modeling of users \cite{Wahlster1986DialogBasedUM, Kobsa2005UserMI, iz2018usermodeling} has been extensively studied within standard dialog modeling framework.
To tackle the difficulty of generating  meaningful responses in an open-domain conversation scenario, many existing models \cite{Walker2012AnAC} implement dialog-based user modeling by generalizing character style on the basis of qualitative statistical analysis.

In contrast, \cite{LiPersona} proposed to train persona vectors directly from conversational data and relevant side-information in a \seqseq~ framework, and incorporate these directly into the LSTM.
While sufficient amount of dialogs with speaker labels are required for these persona-based models \cite{LiPersona, Saizheng2018Personalizing}, they cannot be directly aggregated since personal dialog contains privacy-sensitive information.

% unlabeled open-domain dialog corpus and speaker monologues are leveraged as a substitution \cite{diwang2017outputstyle, yiluan2017speakerrole}, and \cite{Su2019Perdialog} exploit a personalized dialog generation method to reduce requirements of personalized training data.

% Although different data source have some personalized labeled datasets, they cannot be directly aggregated since personal dialog contains privacy-sensitive information. \yj{need dialog privacy reference}
% "Moreover, recent laws and regulations such as GDPR \yj{https:\/\/gdpr-info.eu\/} have enforced strict requirements on protecting the privacy of user data." \yj{need more laws references}

% \yj{add data stream in figure}
\begin{figure*}[h]
    \centering
    \includegraphics[width=\textwidth]{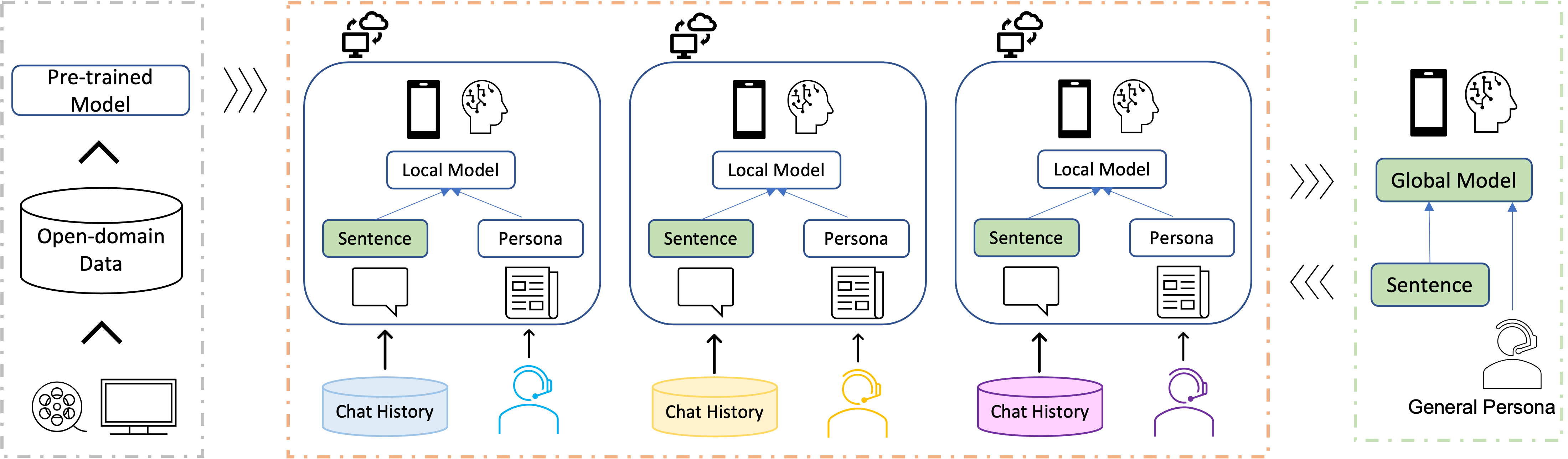}
    \caption{Federated Natural Language Generation Framework. First, a centralized model is pre-trained on open-domain dataset. Then, the pre-trained model is loaded by several local clients. The models fine-tune on personalized dialog corpus locally. Every ${E}$ epoch, the model parameters are aggregated and loaded by central model as one update. The local fine-tuning and aggregation update process is iterated until convergence of central model. Note that only parameters colored in green are globally aggregated. Persona-related parameters are kept locally.}
    \label{fig:FedNLG}
\end{figure*}

\subsection{Federated Learning}
% Each client’s raw data is stored locally and not exchanged or transferred; instead, focused updates intended for immediate aggregation are used to achieve the learning objective.

% applications of user representation learning and 
Federated learning \cite{jakub2016FL, Konecn2016FederatedLS} was first proposed by Google as a decentralized machine learning theory that allows distributed training on a large scale of data in edge-devices such as smartphones, sensors, robots, etc.
% Federated learning is a machine learning setting where multiple entities (clients) collaborate in solving a machine learning problem, under the coordination of a central server or service provider. 
Recent success of federated learning technique in the application of recent increased focus on privacy has earned a remarkable reputation in many pragmatic fields. There are many proposed federated learning framework that extremely boost the development of the these pragmatic fields, such as \cite{Liu2020FederatedLF} for vision-and-language grounding problems, \cite{Kim2021FederatedLF} for face recognition, \cite{Zhang2020BlockchainbasedFL} for device failures detection in IIoT and \cite{Brisimi2018FederatedLO, Kaissis2020SecurePA} for sensitive records analysis. 
Accordingly, the aggregation algorithm such as Federated Averaging \cite{mcmahan2017fedavg} has been widely adopted in practical applications. FedProx \cite{anit2018fedprox} was proposed to tackle heterogeneity in federated networks.
% Some extension works, calculated mutual information of two worker models and suggest the averaging method (\cite{Uddin2021MutualID}), or used a Bayesian non-parametric approach that allows the model to generate expressive matching weights for training Multi-layer Perceptron network on image datasets (\cite{Yurochkin2019BayesianNF}).
% Additionally, communication efficiency is a key challenge of federated learning at present. Some recent works optimized it by low-rank gradient compression for rapid aggregation (\cite{thi2019powersgd}) or applying momentum correction for gradient compression (\cite{Sattler2019SparseBC}).

% federated NLG
Federated learning has also been applied to some natural language processing tasks to exploit the corpus from different sources in a privacy-preserving way \cite{jiang2019federatedTopic, HuangFedSLU, HardFedKeyboard, Chen2019FedNgram, LiuFedVL, Stremmel2020FedText}. However, the application to neural conversational modeling is under explored. Our framework is devised to first incorporate persona-based neural conversational modeling with federated learning techniques, successfully satisfying the laws and regulations without losing user personalized experience.

\section{Methodology}

In this section, we first formulate our problem as persona-based natural language generation. Then we focus on the basic modules of our proposed federated architecture.
% As shown in Figure~\ref{fig:FedNLG}, our proposed federated natural language generation framework (FedNLG) is composed of a persona-based LSTM module which model conversations with consideration of personal information in a sequence-to-sequence manner and a federated module which ensure the data-privacy when utilizing character identification and private dialog history.

\subsection{Problem Formulation}
For the general response generation task, given input word sequence ${Q = \{q_1,q_2,...,q_{L_Q}\}}$, the model aims at predicting the response sequence ${R = \{r_1,r_2,...,r_{L_R}\}}$.
${L_Q}$ and ${L_R}$ is the length of the message and the response respectively, and ${q_t}$ and ${r_t}$ denotes a word token that is associated with a ${K}$ dimensional distinct word embedding ${w_t}$.
In the task of persona-aware conversational generation, the character label ${i}$ would be added to input to generate speaker embedding ${v_i}$, 
${v_i}$ encodes speaker-specific information such as age, gender, location, education and personal preferences, which is shared across all conversations that involve speaker ${i}$.
Then, the input word sequence for speaker $i$ could be represented as ${Q_i = \{v_i, q_1,q_2,...,q_{L_Q}\}}$
% Thus, it is able to predict consistent and stylistic appropriate response.
% %${R}$ denotes the word sequence in response to message ${Q}$.
% All the generated words is selected from the pre-defined vocabulary ${V}$ and ${|V|}$ is the vocabulary size. The embedding sequence of message ${Q}$ and speaker ${i}$ influences the content and style of the responses simultaneously.

\begin{figure*}[h]
    \centering
    \includegraphics[width=\textwidth]{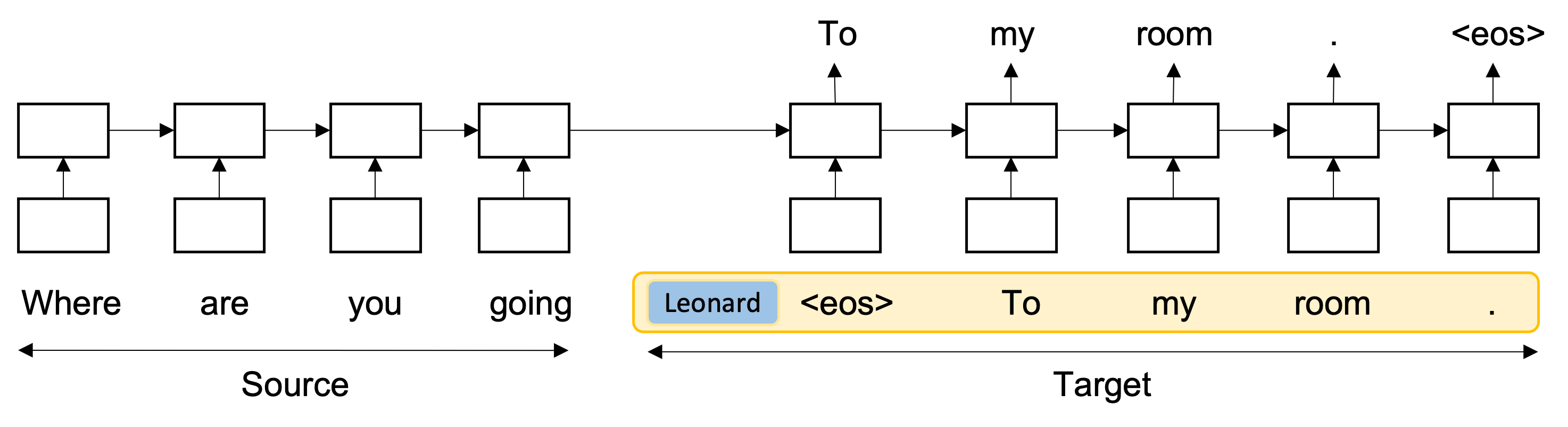}
    \caption{Persona-based sequence-to-sequence framework with federated learning technique. The persona embeddings in blue and word embeddings in yellow are input into the neural model. All the parameters of the embedding layer and neural model are jointly learning. The model learn to predict responses influenced by speaker label, in this example, which is 'Lenard'.}
    \label{fig:seq2seq}
\end{figure*}

\subsection{Persona-embedded Model}
\subsubsection{Encoding}
The model represents each individual speaker as a user-level representation vector ${v_i\in R^{K \times 1}}$, which encodes speaker-specific information (e.g., age, gender, location, education, personal preferences) based on their personal information and dialogue history. The embedding would further shape the style of the responses.

As in standard sequence-to-sequence (\seqseq~) models, we first encode message ${Q}$ into a vector representation ${h_Q}$ using the Recurrent Neural Network (${RNN}$).
Then for each step in the target side, hidden units are obtained by combining the representation produced by the target LSTM at the previous time step, the word representations at the current time step, and the speaker embedding ${v_i}$:

Particularly, we use LSTM cell as the basic ${RNN}$ unit to model dialogue corpus.
Each LSTM unit at time ${t}$ consists of a memory cell ${c_t}$, an input gate ${i_t}$, a forget gate ${f_t}$, and an output gate ${o_t}$. These gates are computed from previous hidden state ${h_{t-1}}$ and the current input ${x_t}$, for standard sequence-to-sequence model:
\begin{equation}
    [f_{t}, i_{t}, o_{t}, j_{t}] = [\sigma, \sigma, \sigma, tanh](W_{seq2seq}[h_{t-1}, x_t])
\label{eq:gates}
\end{equation}
where ${W_{seq2seq} \in R ^ {4K \times 2K}}$ and ${\sigma}$ denotes the sigmoid function. In this way, the model ignore speaker information and thus predicts general responses throughout the generation process.
As for persona-based model:
\begin{equation}
    [f_{t}, i_{t}, o_{t}, j_{t}] = [\sigma, \sigma, \sigma, tanh](W_{persona}[h_{t-1}, x_t, v_i])
\label{eq:gatesPersona}
\end{equation}
where ${W_{persona} \in R ^ {4K \times 3K}}$ and ${\sigma}$ denotes the sigmoid function.. In this way, speaker information is encoded and injected into the hidden layer at each tie step and thus helps predict personalized responses throughout the generation process. The Speaker embedding ${v_i}$ is shared across all conversations that involve speaker ${i}$. ${v_i}$ are learned by back propagating word prediction errors to each neural component during training.
The memory cell ${c_t}$ is updated by partially forgetting the existing memory and adding a new memory content ${j_t}$:
\begin{equation}
    % j_{t} = tanh(V[h_{t-1}, x_t]),
    c_{t} = f_t \odot c_{t-1} + i_t \cdot j_t
\label{eq:predictiona}
\end{equation}
Once the memory content of the LSTM unit is updated, the hidden state at time step ${t}$ is given by:
\begin{equation}
    h_{t} = o_{t} \odot tanh(c_t)
\label{eq:predictionb}
\end{equation}
At time step ${t}$, the new states of the user can be inferred as:
\begin{equation}
    h_{i,t}^{v} = LSTM(h_{i,t-1}^{v}, z_{i,t}^{v})
\label{eq:newState}
\end{equation}
where ${h_{i,t}^{v}}$ denote hidden states for the speaker ${v}$. 

Note that LSTM can be replaced by other options, such as the third version of generative pre-train transformer (GPT-3) proposed in \cite{Radford2018ImprovingLU} and a state-of-the-art large-scale pre-trained dialogue response generation model (DialoGPT) proposed in \cite{Zhang2020DIALOGPTL}, etc. 
In this paper, we focus on exploiting federated learning into natural language generation (NLG) task. For simplicity, we select LSTM to evaluate the effectiveness of our federated framework to address two major challenges, consistency with humanity and privacy-accuracy balance. %Furthermore, LSTM is lightweight, which is good for focusing our experiments on the investigation of federated personalized modules in this scenario.

\subsubsection{Decoding}
For decoding, we adopt the log-likelihood of target response ${R}$ given source question ${Q}$ as our objective function for \methodname~ as follows:
\begin{equation}
    \hat{R} = \arg\max_{R} {\log{p(R|Q)}}
\end{equation}
We conduct beam search over the above objective with beam size ${B = 200}$ and a maximum length of the generated candidates ${20}$ to generate the $N$-best lists. 
% At each time step, we first examine all ${B \times B}$ possible next-word candidates, and add all hypothesis ending with an ${EOS}$ token to the N-best list. We then preserve the top-B unfinished hypotheses and move to the next word position.

\subsubsection{Reranking}
We follow \cite{Li2016ADO} by adopting a scoring function for reranking the generated $N$-best list.
The scoring function linearly combines a length penalty and the log likelihood of the source question given the target response as follows:
\begin{equation}
    \log{p(R|Q, i)} + \lambda \log{p(Q|R)} + \gamma |R|
\end{equation}
where ${p(R|Q,i)}$ denotes the probability of generating response $R$ given the source question ${Q}$ and the respondent's speaker identification ${i}$.
${|R|}$ denotes the length of the target response and ${\gamma}$ denotes the associated penalty weight.
We optimize ${\gamma}$ and ${\lambda}$ using Minimum Error Rate Training \cite{Och2003MinimumER} by optimizing BLEU.
An inverse standard \seqseq~ model with no speaker information was trained to compute ${p(Q|R)}$.

\subsection{Federated Framework}
% The main constraint in preserving privacy while learning user embeddings is that user data should not be transferred back to the server nor distributed to any other users.
%While typical model parameters are trained on data from all users, user embeddings are very privacy-sensitive because a user's embedding is trained only on that user's data.

% intuition of parameter splitting for federated personalized NLG
In order to guarantee the model performance, data-privacy, and preserve personalization, we empirically found that simply applying Federated Averaging or any other federated algorithms in neural conversational model would achieve poor performance in persona learning.
We suspect it is due to the global aggregation of persona parameters which lead persona to general representation and fail to fully exploit local personal data.
Thus, to enable models to learn personal representation, we first separate our learning parameters in the framework into tow parts, \textit{federated} part and \textit{private} part.
For the word embeddings and neural model parameters, we categorize them into the ${federated}$ part, while for the persona embeddings ${v_i}$, we regard them as ${private}$ part.

% Our model use common personalization techniques, personalization via embeddings, which satisfy the ${split-personalization}$ constraint when trained using ${Federated Averaging}$.
Our framework \methodname~ include two phases: 1) open-domain dialog pre-training; and 2) persona-based fine-tuning under federated frameworks.
Federated fine-tuning phase is composed of local training and federated aggregation.

\textbf{Pre-training} To obtain a general dialog model, we pre-train a standard \seqseq~ model by optimizing ${BLEU}$ scores on publicly available dataset.
In our expeirments, we utilize \cornell~ Corpus.

\textbf{Local Training} 
%\tbbt~ and \friend~ are distributed to the specific client according to conversation utterances' speaker identifications.
Each client's conversation utterances will be trained by separated models.
Each model locally fine-tunes based on its assigned character's private conversation history.
Neither do central server have assess to local clients' data nor do local clients have access to any other characters' data.
At the start of local fine-tuning, the model of character ${i}$ uniformly initializes persona parameters as ${v^{i,t}_{p,i}:=w^{i,t}_{p}}$ and load other parameters from pre-trained model as ${v^{t}_{f,i}:=w^{t}_{f}}$. ${v}$ represents a local parameter that will change during local training, ${v^{i,t}_{p,i}}$ denotes persona parameters of user ${i}$ stored locally on user ${i}$'s device. In each iteration ${t}$, local model optimize with gradient descent on the model of user ${i}$, which updates its federated parameters ${v^{t}_{f,i}}$ and personal parameters ${v^{i,t}_{p,i}}$.

\textbf{Federated Aggregation} When local fine-tuning reaches certain number of epochs, federated aggregation begins.
In particular, locally updated federated parameters ${v^{t}_{f,i}}$ are sent to the server for global aggregation while personal parameters ${v^{i,t}_{p,i}}$ are still kept in local model.

% ${w^{i,t+1}_{p,i}}$.

% \yj{may use original theory provement? is this work well recognized? maybe we should not mention it or rely on it as our theory basis?}
As discussed in \cite{BuiFedUserRepresentation}, parameters splitting is valid for any FL algorithms that satisfy Equation~\ref{equ:splitParamFed} for federated parameters update and Equation~\ref{equ:splitParamPrivate} for private parameters update.
\begin{equation}
    w^{t+1}_{f} := a_{f} (w_{f}^{t}, v_{f,1}^{t},..., v_{f,n}^{t}, s_1, ..., s_n)
\label{equ:splitParamFed}
\end{equation}

\begin{equation}
    w^{i, t+1}_{p} := a_{p} (w_{p}^{i,t}, v_{p,i}^{i,t}, s_1, ..., s_n)
\label{equ:splitParamPrivate}
\end{equation}
where the function ${a_f}$ represents federated aggregation function, while the function ${a_p}$ is the identity function. ${s_{i}}$ represents summary information about data of character ${i}$.

% details of federated variants
Our proposed \methodname~ satisfies such constraint by assigning persona-related parameters as private parameters and other parameters as federated parameters.
% in a single task manner. multi-task learning may be a bottleneck
Specifically, we train our neural conversational model to predict responses given questions and adopt configuration of commonly used federated aggregation algorithms, such as Federated Averaging \cite{mcmahan2017fedavg} and Federated Proximal \cite{anit2018fedprox}. For FedAvg Variant, the update of federated parameters, including word embeddings and neural model parameters is as follows:
\begin{equation}
    w^{t+1} = \frac{ \sum_{i=1}^{n}(d_{i}^{t} + w^{t})c_{i} } { \sum_{i=1}^{n}c_{i} } = w^{t} + \frac{ \sum_{i=1}^{n}(d_{i}^{t})c_{i} }{ \sum_{i=1}^{n}c_{i} }
\end{equation}
As persona parameters are kept in local client, the update equation is as follows:
\begin{equation}
    w^{i, t+1}_{p} = w^{i, t}_{p} + \frac{ \sum_{j=1}^{n}d_{p,i}^{j,t}c_{j} }{ \sum_{i=1}^{n}c_{i} }
    % = w_{p}^{i,t} + z_{i}d_{p,i}^{i,t}
\end{equation}
where ${c}$ denotes the training samples for character ${i}$, ${d}$ denotes model delta of parameters of character${i}$, and ${n}$ denotes the total number of characters. For ${\forall j \ne i}$, ${d_{p,i}^{j,t}}$ is set to ${0}$ to guarantee that no clients persona parameters are distributed to any other clients or central server.

\section{Experiments}
In this section, we first introduce the data used in experiments, and evaluation metrics. 
Then, we introduce the methods to compare, and conduct extensive experiments to demonstrate the effectiveness of the proposed framework.

\begin{table*}[h]
\centering
    \caption{Overall Performance Comparison. The model performance are evaluated in BLEU scores and Perplexity on \tbbt~ and \friend~. \seqseq~ (+FL) is a standard \seqseq~ implemented in our framework with observed best configuration, while \seqseq~(+Persona) is a persona-based model proposed in \cite{LiPersona}. Ours (\methodname~) is implemented in ${3}$ federated configurations. The best results are highlighted in bold and the second best results are underlined.}
\resizebox{\textwidth}{!}{%    
\begin{tabular}{l l ll ll ll}
\toprule
&&\multicolumn{2}{c}{\textbf{The Big Bang Theory}} & \multicolumn{2}{c}{\textbf{Friends}} \\
\cmidrule(lr){3-4}\cmidrule(lr){5-6}\
\#&\textbf{Model} & BLEU Score $\uparrow$ & Perplexity $\downarrow$    & BLEU Score $\uparrow$ & Perplexity $\downarrow$ \\
    \midrule
    % Base Model (SEQ2SEQ)      & 0.142 & 0.030 & 0.007 & ${\mathbf{56.3}}$ & 0.135 & 0.030 & 0.003 & ${\mathbf{32.4}}$\\
    0 &SEQ2SEQ            & 0.108 & 65.0 & 0.116 & 34.7\\
    1&~~~ $+$ FL               & ${0.099 (-8.3\%)}$ & ${67.0 (+3.1\%) }$ & ${0.102(-12.1\%)}$ & ${47.3(+36.3\%)}$\\
    2&~~~ $+$ Persona          & ${\mathbf{0.131} (+21.3\%)}$ & ${52.0 (-20.0\%)}$ & ${\mathbf{0.137} (+18.1\%)}$ & ${59.0 (+70.0\%)}$\\ 
    \midrule
    3&FedNLG (FedProx)            & ${\underline{0.125} (+15.7\%)}$ & ${\mathbf{37.8} (-41.8\%)}$ & ${\underline{0.133} (+14.7\%)}$ & ${\mathbf{28.5} (-17.9\%)}$\\
    4&FedNLG (FedDrop)            & ${0.100(-7.4\%)}$ & ${\underline{39.6}(-39.1\%)}$ & ${0.096(-17.2\%)}$ & ${\underline{29.2} (-15.9\%)}$\\
    % FedNLG(Sample-Weighted)   & ${0.098(+xx\%)}$ & ${67.3(+xx\%)}$ & ${0.138(+xx\%)}$ & ${51.7(+xx\%)}$ \\
    5&FedNLG (FedAvg)             & ${0.111(+2.8\%)}$ & ${62.1(-4.5\%)}$ & ${0.130 (+12.1\%) }$ & ${52.4 (+51.0\%)}$\\
    % FedNLG (fix local persona weight FL)     &  &  &  & \\
    % FedNLG (persona weight FL)  &  & & 0.086 & 33\\
    % FedNLG (all FL)          &  & & 0.086 & 33\\
    \bottomrule
\end{tabular}
}
    \label{tab:modPer}
\end{table*}

% \begin{table*}[h]
% \centering
%     \caption{Performance Comparison of Federated Module Variants. The model performance of several federated module variants of our model on \tbbt~ and \friend~. The best results are highlighted in bold.}
% \resizebox{\textwidth}{!}{%    
% \begin{tabular}{l cc cc cc cc cc cc}
% \toprule
% &\multicolumn{4}{c}{\textbf{The Big Bang Theory}} & \multicolumn{4}{c}{\textbf{Friends}} \\
% \cmidrule(lr){2-5}\cmidrule(lr){6-9}\
% \textbf{Model} & BLEU-1 & BLEU-2 & BLEU-3 & Perplexity & BLEU-1 & BLEU-2 & BLEU-3 & Perplexity \\
%     \midrule
%     FedAvg                 & ${\mathbf{0.111}}$ & 0.017 & 0.003 & 62.1 & ${\mathbf{0.136}}$ & 0.028 & 0.006 & 47.3\\
%     Sample-Weighted Averaging & 0.098 & 0.019 & 0.002 & 67.3 & 0.138 & 0.026 & 0.0006 & 51.7 \\ 
%     RandomDrop             & 0.100 & ${\mathbf{0.027}}$ & 0.003 & 39.6 & 0.096 & 0.018 & 0.003 & 29.2\\
%     FedProx                & 0.125 & ${\mathbf{0.027}}$ & ${\mathbf{0.005}}$ & ${\mathbf{37.8}}$ & 0.102 & ${\mathbf{0.029}}$ & ${\mathbf{0.004}}$ & ${\mathbf{27.3}}$\\
%     \bottomrule
% \end{tabular}
% }
%     \label{tab:fedVariant}
% \end{table*}

\subsection{Data}
%We use two types of dataset to train and evaluate our model, one for open-domain dialogue pre-train and another for persona-based dialogue fine-tune.

% All three open datasets are from English Movie and TV series.
% The details of each dataset are described as below:
% \begin{itemize}
% \item {\verb|OpenSubtitles|\cite{SuPerDialogue}}: "OpenSubtitles dataset1 [22] consists of movie subtitles and is used to provide a general conversational knowledge and to keep the contextual coherence in our experiment. Since OpenSubtitles is a large and noisy corpus [23] containing a good amount of generic sentences like ’I don’t know’ and ’I’m sorry’, we filtered out those inputresponse pairs containing the frequent generic responses with a defined threshold. The pruned training set has in total 453, 106 lines."
% \item {\verb|BigBang|\cite{SuPerDialogue}}: TBBT. "To compare with prior persona models [1], we used dialogues in American TV series Friends and The Big Bang Theory (TBBT). "
% \item {\verb|Friends|\cite{SuPerDialogue}}: Friends. "To compare with prior persona models [1], we used dialogues in American TV series Friends and The Big Bang Theory (TBBT). "
% \end{itemize}
\begin{itemize}
\item \textbf{Cornell Movie-Dialogs} \cite{cristian2011cornell} contains ${220,579}$ conversational messages between ${10,292}$ pairs of movie characters, which involves ${9,035}$ characters from ${617}$ movies. We ignore the character meta-data information and use this corpus as general movie dataset to pre-train our open-domain dialogue model.

\item \textbf{The Big Bang Theory:} To compare with prior persona models, we use dialogues in American TV series The Big Bang Theory to fine-tune our pre-trained model. It contains ${1,075}$ characters and total ${51,473}$ utterances. 
We collect ${7}$ main characters from this series.
We randomly split the corpus into training, development and testing set with ratio $80\%$, $10\%$ and $10\%$ respectively.
% ['sheldon','leonard','penny','howard','amy','bernadette','cody'] 9040,7602,5940,4540,2720,2100

\item \textbf{Friends:} To compare with prior persona models, we use dialogues in American TV series Friends to fine-tune our pre-trained model. It contains ${962}$ characters and total ${63105}$ utterances. We collect ${6}$ main characters from this series.
We randomly split the corpus into training (70\% of utterances), development (10\%) and testing set (10\%), respectively.
% rachel 7488,ross 7473,chandler 6853 monica 6806 joey 6723 phoebe 6060
\end{itemize}

\subsection{Evaluation Metrics}
To compare the performance of different models, we use the following 2 metrics:
\begin{itemize}
\item {\verb|BLEU|}: We use BLEU scores to fine-tune and evaluate models. It is an accuracy measure quantifying the precision of the generated response. 
\item {\verb|Perplexity|}: We use perplexity to evaluate the capability of models, which describe the degree of surprise of a model. The perplexity is inversely correlated to the precision of the probability that model predicts of responses.
\end{itemize}

\subsection{Experimental Configuration}
\label{sec:exp_setting}
We pre-train four-layer \seqseq~ base model on \cornell~ Corpus without access to their character metadata. Then we fine-tine on the \friend~ and \tbbt~ dataset. Details are as follows:
\begin{itemize}
    \item Vocabulary size is limited to ${30,000}$. Each sentence length is limited under ${20}$.
    \item The LSTM is ${4}$ layer with ${100}$ hidden cells for each layer. The dimension of persona embedding is set to ${128}$.    
    \item Batch size is set to ${1024}$. Learning rate is set to ${0.01}$. Dropout rate is set to ${0.2}$. Gradients are clipped to avoid gradient explosion with a threshold of ${5}$.
    % sampling from the uniform distribution ${[−0.1, 0.1]}$
    \item Parameters are initialized with Xavier uniform initializer when pre-training. During fine-tuning, the persona embedding layer is initialized to normal distribution, while other parameters are loaded from pre-trained model.
    \item We use 60 epochs on the pre-training and 90 epochs for the federated phase.
\end{itemize}

%\subsection{Model Training}
%First we pre-train our model on open-domain dataset without considering personalization.
%Then we fine-tune our model in a federated manner.
%Fine-tuning Process are divided into two steps, local training and federated aggregation.

% \begin{itemize}
% \item {\verb|Local Training|}: Local model are trained with only access to their local data.
% \item {\verb|Global Aggregation|}: At the end of local training, locally updated persona parameters are aggregated.
% \end{itemize}

%\methodname~  pre-train by using Stochastic Gradient Descent and fine-tune by combining SGD from selected client models with a federated module configuration, for instance, Federated Averaging.
%The model were pre-trained for ${60}$ epochs on \cornell~ and fine-tuned for ${90}$ on \tbbt~ and \friend~.

\begin{table*}[h]
\centering
    \caption{Personalized responses. The personalized responses generated for \tbbt~ and \friend~ by our proposed \methodname~ with FedProx configuration. The responses of standard \seqseq~ and \persona~ model are reported for reference.}
\resizebox{\textwidth}{!}{%    
\begin{tabular}{l cc cc cc cc cc cc}
\toprule
&\multicolumn{4}{c}{\textbf{The Big Bang Theory}} & \multicolumn{4}{c}{\textbf{Friends}} \\
\cmidrule(lr){2-5}\cmidrule(lr){6-9}\
\textbf{Person} & \multicolumn{2}{c}{Sheldon} & \multicolumn{2}{c}{Leonard} & \multicolumn{2}{c}{Rachel} & \multicolumn{2}{c}{Ross} \\
\midrule
% \cmidrule(lr){2-3}\cmidrule(lr){4-5}\cmidrule(lr){6-7}\cmidrule(lr){8-9}\
\textbf{Question} & \multicolumn{4}{c}{Well, this calls for an expression of gratitude.} & \multicolumn{4}{c}{What if i smack my head on the concrete?} \\
    \midrule
    \seqseq~ & \multicolumn{4}{c}{No, I have a little more.} & \multicolumn{4}{c}{Well, I just wanted to talk to you.}\\ 
    \cmidrule(lr){2-3}\cmidrule(lr){4-5}\cmidrule(lr){6-7}\cmidrule(lr){8-9}\
    \persona~ & \multicolumn{2}{c}{So, what do you say?} & \multicolumn{2}{c}{So, how'd it go? } &
    \multicolumn{2}{c}{Hey, you guys, you're not gonna be able.} & \multicolumn{2}{c}{What are you doing?}\\ 
    \cmidrule(lr){2-3}\cmidrule(lr){4-5}\cmidrule(lr){6-7}\cmidrule(lr){8-9}\
    \methodname~  & \multicolumn{2}{c}{He's not gonna get married.} & \multicolumn{2}{c}{He's not a little worried about you.} &
    \multicolumn{2}{c}{You're not gonna be a girl and Monica?} & \multicolumn{2}{c}{You're not gonna be a genius?}\\ 
\midrule    
\textbf{Question} & \multicolumn{4}{c}{Your love confuses me.} & \multicolumn{4}{c}{Come on, what, you never think about our future?} \\
    \midrule
    \seqseq~ & \multicolumn{4}{c}{Yeah. I mean, like the same thing.} & \multicolumn{4}{c}{Oh , I'm sorry. }\\ 
    \cmidrule(lr){2-3}\cmidrule(lr){4-5}\cmidrule(lr){6-7}\cmidrule(lr){8-9}\
    \persona~ & \multicolumn{2}{c}{That's a good idea.} & \multicolumn{2}{c}{Do you know what I do?} &
    \multicolumn{2}{c}{I'm sorry, I'm sorry.} & \multicolumn{2}{c}{I'm sorry.}\\ 
    \cmidrule(lr){2-3}\cmidrule(lr){4-5}\cmidrule(lr){6-7}\cmidrule(lr){8-9}\
    \methodname~  & \multicolumn{2}{c}{So, I don't know.} & \multicolumn{2}{c}{Why?} &
    \multicolumn{2}{c}{Oh, okay... What?!} & \multicolumn{2}{c}{Oh, okay... And Monica, I'm sorry, I'm sorry.}\\
    \bottomrule
\end{tabular}
}
    \label{tab:personaCase}
\end{table*}
% the point is i . . . i don t need this right now

\subsection{Quantitative Comparison to the State-of-the-Art Methods}
To validate the effectiveness of our proposed \methodname~ framework, we compare the proposed framework with the following 3 methods on two publicly available TV series datasets, \tbbt~ and \friend~:
%The comparison to a standard sequence-to-sequence model and a person-based model are summarized in Table~\ref{tab:modPer}. For fair comparison, all the models are implemented in the same experimental settings.
\begin{itemize}
% Previous conversational models are mainly rule-based. 
\item {\verb|SEQ2SEQ|}: This work \cite{Vinyals2015neural} proposed a neural conversational model, which makes use of the sequence-to-sequence framework, which is based on a recurrent neural network.
\item {\verb|Persona-based Model|}: \cite{LiPersona} present persona-based model to address the challenge of producing responses with consistency and persona in conversational model. They explore to embed persona into an LSTM to produce personalized responses. This model is implemented with full access to character metadata and thus face the risk of private data leakage.
\item {\verb|SEQ2SEQ with Federated Learning|}: We also implement the standard SEQ2SEQ model in federated framework as a base model without access to character metadata, thus no personalization.
% \item {\verb|Ours(FedAvg)|}: 
% \item {\verb|Ours(FedWeightedAvg)|}:
% \item {\verb|Ours (FedProx)|}: 
\end{itemize}

% Base Model Comparison
We report performance our proposed \methodname~ on \tbbt~ and \friend~ dataset over metric of BLEU scores and Perplexity are reported in Table~\ref{tab:modPer}. Experimental results for Standard \seqseq~, \seqseq~(+FL) and \seqseq~ (+Persona) are reported for reference.

As can be seen in Table~\ref{tab:modPer}, for \tbbt~ and \friend~ TV series dataset, \methodname~ (FedProx) improve significant relative performance than standard \seqseq~ model up to ${15.7\%}$ and ${14.7\%}$ in BLEU scores, ${41.8\%}$ and ${17.9\%}$ in perplexity, respectively.
The improvement indicates effectiveness of our proposed \methodname~ to incorporate personalized natural language generation with federated learning techniques.
\methodname~ achieves similar results over BLEU scores and perplexity as \persona~ model. The major difference between these two models is that \persona~ is implemented with full access to global data which also lead to data privacy issue, while \methodname~ only access local client data. Thus the small drop in metrics compared to global \persona~ is expected and tolerable, which is also discussed in \cite{mcmahan2017fedavg}. This comparison result indicates effectiveness of our proposed model to tackle the challenge of balancing between model accuracy and data privacy.
% A little drop performance in metrics than globally trained \persona~ model is tolerable and reasonable, as it is intuitive that the federated fine-tuning process have no access to global character information and dialog corpus.
Similarly, we observe small decrease in BLEU scores for the \seqseq~ (+FL) compared to the standard \seqseq~. Besides, we observe small decrease in the perplexity, which is not uniform with BLEU performance. We suspect it is due to that the model fluctuates with predictions to similar options, as no persona is utilized for \seqseq~ (+FL), while still gets more confident in cases of obvious difference. This observation is also consistent with research work \cite{rikters2016neural}.
% This indicates that out of very similar options, even the NN model fluctuates with its predictions but it does get more confident in cases where the difference is more obvious.

\subsection{Comparison to Different Federated Learning Algorithms}
As the proposed framework is ubiquitous for various federated learning algorithms, we further analyze the performance of using different federated technique under our proposed model and report the experimental results in Table~\ref{tab:modPer}.
We consider the following federated learning algorithms in the following experiments: 
%We train variants with following configurations of our proposed flexible \methodname~, which could adaptively incorporate existing common federated algorithms:
\begin{itemize}
    \item {\verb|FedAvg|} \cite{mcmahan2017fedavg} : We adopt Federated Averaging to implement a federated module variant. This variant could learn parameters based on iterative model averaging between global server and clients.
    \item {\verb|FedDrop| \cite{Ji2020DynamicSA}}: We add some modification to original FedAvg by different client drop out strategy in each aggregation round for communication-efficiant federated variant.
    \item {\verb|FedProx|} \cite{anit2018fedprox}: We adopt Federated Proximal , which can be viewed as a generalization and re-parametrization of FedAvg \cite{mcmahan2017fedavg}. We implement this federated module variant to compete with other federated modules.
    % \item {\verb|Sample Weighted Averaging|}: We add some modification to original \cite{mcmahan2017fedavg} by positive correlate the sum weight of each client to the number of its samples. This variant suffers from im-balanced samples of different clients.
\end{itemize}

% Not surprisingly, \methodname~ successfully incorporate persona-related parameters into federated fine-tuning, as its performance is much more better than the base model \seqseq~.
% federated parameters seperation
As shown in Row 1 in Table~\ref{tab:modPer}, we empirically find that simply applying federated techniques to neural conversational models not only fails to learn persona representation, but also results in poor performance and producing less diversified responses.
Motivated by some research work established in \cite{BuiFedUserRepresentation}, we devise our framework to be able to adaptively separate parameters into persona part to learn persona embedding locally, and federated part, which could be aggregated globally to train neural network.
% federated configuration
Furthermore, we establish several variants of our federated modules to investigate the effectiveness of different federated configurations. Performance of different configurations of our framework, including FedDrop, FedAvg ,and FedProx are reported in Table~\ref{tab:modPer}.
Not surprisingly, \methodname~ (FedDrop) achieves poor performance due to it uses less information from clients.
%Somehow, it also indicates that other variants could learn such local information during aggregation without access to local data.
\methodname~ (FedAvg) achieves relatively good performance in both TV series dataset, which is consistent with it's claimed to cope with imbalanced data and not independent and identically distributed (non-iid) data. 
It is worth noting that \methodname~ (FedAvg) achieves relatively bad performance on \tbbt~ compared to \friend~. This result is intuitive as \tbbt~ is more heterogeneous than \friend~, with more variance in characters lines and more diversified conversational context with differed distribution for different characters. 
% Sample Weighted Averaging is a modified implementation of FedAvg by simply change sum to weighted sum according to number of each client's samples. As samples are imbalanced in both \tbbt~ and \friend~ dataset, it's no doubt that we observe a drop in metric.

\methodname~ (FedProx) achieves best performance over both BLEU and perplexity on \tbbt~ and \friend~ under our framework.
It also achieve better performance than \methodname~ (FedAvg). We suspect this is primarily due to that our dataset is not independent and identically distributed (non-iid), thus \methodname~ (FedProx) has some advantage over \methodname~ (FedAvg) as expected, which is also consistent with previous research work \cite{anit2018fedprox}.
In addition, personalized federated learning seeks to reduce heterogeneity and maintain high-quality client contributions to central server as discussed in \cite{jiehan2021survey}, which is successfully by our FedProx configuration.
Thus we demonstrate that FedProx with separated parameters techniques is a state-of-the-art configuration in our proposed framework, and probably in the scenario of federated personalized natural language generation.
These results also demonstrate the flexibility of our framework to incorporate existing widely used federated algorithms.

\subsection{Qualitative Comparison to the State-of-the-Art Methods}
To further illustrate the effectiveness of personalized conversational response generation, we report random selected cases in Table~\ref{tab:personaCase}. The \seqseq~ represent the standard sequence-to-sequence model, \persona~ represent the persona-based Speaker Model proposed in \cite{LiPersona}, and \methodname~ represent personalized conversational model with federated learning technique.
As can be seen, the standard \seqseq~ produces the same response to the given question for different characters, which is intuitive that no persona was learned in this model. 
In contrast, we observe that \persona~ model and our proposed \methodname~ are both sensitive to the identity of the character, generating diversified responses which contains persona and humanity. For example, the model produces "You're not gonna be a genius?" in response to "What if i smack my head on the concrete?", which could be recognized as "Ross" style.
Both \persona~ and \methodname~ predicts diversified responses with persona in two movie datasets, \friend~ and \tbbt~.
We also test our model on consistency by generating some similar questions, empirically our persona-based models successfully produce consistent responses compared to standard \seqseq~ model, which is consistent with results in \cite{LiPersona}.
% Not surprisingly, the \methodname~ without persona failed to generate responses with consistency or personality.

\section{Conclusion}
We introduce a framework, Federated Natural Language Generation (\methodname~), to tackle two major challenges in personalized natural language generation, producing consistent responses with humanity and achieving accuracy-privacy balance.
\methodname~ is a novel flexible framework that could learn the personalized representations from various dataset on distributed devices, and thus implement the personalized neural conversational model efficiently and safely.
% \methodname~ first pre-trains general embedding over everyday conversation dialogs, \cornell~ Corpus, and then fine-tune the persona embedding over domain-specific dataset, \tbbt~ and \friend~ in a federated manner. 
As both quantitative and qualitative results show, \methodname~ significantly improves performance over non-personalized standard \seqseq~ model and achieve similar performance to global trained \persona~ model with superiority of accuracy-privacy balance. Investigation on different federated variants also indicates our proposed framework is flexible enough to incorporate with existing popular federated techniques. 
% Extensive experiments show that our proposed \methodname~ could efficiently learn the persona embedding and background information in local clients and incorporate general embedding simultaneously.

\newpage
\bibliography{main.bib}

\end{document}